%% file: cxrnaacl.tex
\documentclass[11pt,a4paper]{article}
\usepackage{amssymb,amsmath,amsthm,mathtools,url,microtype,xcolor,epsfig}
\usepackage[]{naaclhlt2019} %% hyperref
\usepackage{times}
\usepackage{latexsym}
\usepackage{url}
\usepackage{float}

\aclfinalcopy % Uncomment this line for the final submission
\def\aclpaperid{123} %  Enter the acl Paper ID here
\title{Scaling Multi-Domain Dialogue State Tracking via Query Reformulation}
\author{
Pushpendre Rastogi \thanks{Work done while the author was at Alexa AI} \\ {\tt pushpendre@jhu.edu} \\Johns Hopkins University\\ Baltimore, MD,USA \\\And
Arpit Gupta \\ {\tt arpgup@amazon.com} \\Alexa AI \\Amazon.com, Inc., USA  \\\AND
Tongfei Chen$^{*}$\\ {\tt tongfei@jhu.edu} \\Johns Hopkins University \\Baltimore, MD,USA \\\And
Lambert Mathias\\ {\tt mathiasl@amazon.com} \\Alexa AI \\Amazon.com, Inc., USA }
\newcommand{\ath}[1]{#1^{\textrm{th}}}
\newcommand{\figref}[1]{Figure~\ref{#1}}
\newcommand{\tabref}[1]{Table~\ref{#1}}

\newcommand{\secref}[1]{Section~\ref{#1}}

\newcommand{\dlx}[3]{$\frac{\text{\textit{Entity}#1:{\color{darkgreen}#2}}}{\text{\color{red}#3}}$}
\definecolor{darkgreen}{rgb}{0.0, 0.5, 0.13}
\newcommand{\bh}{\mathbf{h}}
\newcommand{\x}{\mathbf{x}}
\newcommand{\y}{\mathbf{y}}

\usepackage{enumitem}
\begin{document}
\maketitle
\begin{abstract}
We present a novel approach to dialogue state tracking and referring expression resolution tasks. Successful contextual understanding of multi-turn spoken dialogues requires resolving referring expressions across turns and tracking the entities relevant to the conversation across turns. Tracking conversational state is particularly challenging in a multi-domain scenario when there exist multiple spoken language understanding (SLU) sub-systems, and each SLU sub-system operates on its domain-specific meaning representation. While previous approaches have addressed the disparate schema issue by learning candidate transformations of the meaning representation, in this paper,  we instead model the reference resolution as a dialogue context-aware user query reformulation task~\textemdash the dialog state is serialized to a sequence of natural language tokens representing the conversation. We develop our model for query reformulation using a pointer-generator network and a novel multi-task learning setup. In our experiments, we show a significant improvement in absolute F1 on an internal as well as a, soon to be released, public benchmark respectively.
\end{abstract}

\section{Introduction}
\label{sec:introduction}
\begin{figure}[htbp]
  \centering
  \includegraphics[width=\linewidth]{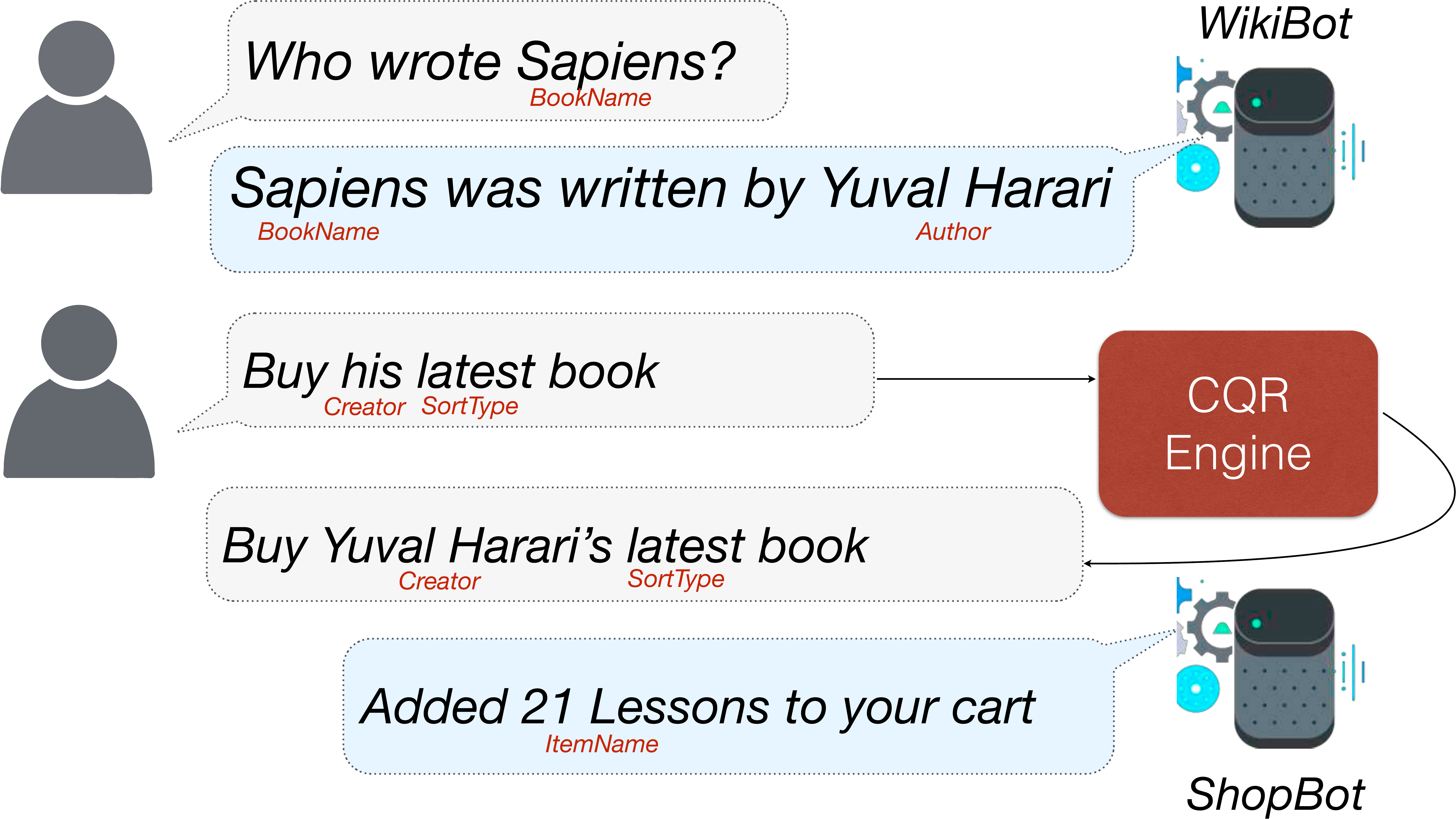}
  \caption{An example dialog where the second utterance by the user \textsc{Buy his latest book} is reformulated as \textsc{Buy Yuval Harari 's latest book}. This reformulated user query is then input to \textsc{ShopBot} so that it can understand the user's request using its existing SLU logic for handling single-turn queries. This approach does not require any changes to the agent itself and can be scaled to multiple heterogeneous domains.}
  \label{fig:dialog1}
\end{figure}

Dialogue \textit{assistants} are used by millions of people today to fulfill a variety of tasks. Such assistants also serve as a digital marketplace\footnote{https://dialogflow.com} \cite{kumar2017just} where any developer can build a domain-specific, task-oriented, dialogue \textit{agent} offering a service such as booking cabs, ordering food, listening to music, shopping etc. Also, these agents may interact with each other, when completing a task on behalf of the user. \figref{fig:dialog1} shows one such interaction where the agent -- ShopBot -- must interpret the output of the agent -- WikiBot.
% Put image in middle of para so that it renders on the top of the second column on first page.
Often accomplishing this task requires understanding the context of a dialogue, communicating the conversational state to multiple agents and updating the state as the conversation proceeds.

Tracking the dialogue state across multiple agents is challenging because agents are typically built for single-turn experiences, and must be laboriously updated to handle the context provided by other agents into their respective domain specific meaning representation. \cite{naik2018contextual} proposed \textit{context carryover}, a scalable approach to handle disparate schemas by learning mappings across the meaning representations, thereby eliminating the need to update the agents. However, the challenge of the agent's domain-specific SLU accuracy and choice of meaning representation remains. For example, in \figref{fig:dialog1} the \textsc{ShopBot} cannot handle pronominal anaphora and instead incorrectly labels \textsc{his} as the mention type \textsc{Creator}. Separately solving this problem for each agent, imposes a  burden on the developer to relabel their data and update their SLU models, and is expensive and unscalable. Moreover, this approach cannot leverage the syntactic regularities imposed across agents by the natural language itself.

In this work, we propose a novel approach for enabling seamless interaction between agents developed by different developers by using natural language as the API. We build upon the pointer-generator network (PGN) proposed by~\cite{see2017gttp} -- originally for news article summarization -- to rewrite user utterances and disambiguate them. Furthermore, we describe a new Multi-task Learning (MTL) objective to directly influence the attention of the PGN without requiring any extra manually annotated training data. Our results show that the new MTL objective reduces the error by $3.2\%$ on slots coming from distances $\geq$3, compared to the basic PGN by~\cite{see2017gttp}.

%To summarize, our main contributions are presenting dialogue context based user query rewriting as an alternative to traditional approaches to multi-domain dialogue state tracking and reference resolution. We also show a novel multi-task loss for training pointer-generator networks and demonstrate that it improves the performance of our system.

\section{Technical Details}
\label{sec:method}
\input{methods}

\section{DataSet and Preprocessing}
\label{sec:data}

\input{data}

\section{Experiments}
\label{sec:experiment}
\input{experiments}

\section{Results}
\label{sec:results}
\input{results}

\section{Related Work}
\label{sec:bg}
\input{background}

\section{Conclusion}
\label{sec:conclusion}
\input{conclusion}

\bibliographystyle{acl_natbib}
\bibliography{naaclhlt2019}

\end{document}

%% file: methods.tex
\begin{figure*}[h!]
  \centering
  \begin{tabular}{r | p{13.5cm}}
    \textbf{Input} $x_{t=2}$ & \textsc{BookQuery}$_{l=1}$ Who wrote \dlx{U1}{BookName}{Sapiens}$_{l=4}$ $\frac{\textsc{System}}{\textsc{InformIntent}}$ \dlx{U1}{Sapiens}{Title} was written by \dlx{S1}{Author}{Yuval Harari}$_{l=10}$ $\frac{\textsc{User}}{\textsc{UnkIntent}}$ Buy \dlx{U2}{Entity}{his} \dlx{U3}{Entity}{latest} book \textsc{END}$_{l=16}$\\[5pt]\hline
    \textbf{Refer.} $\mathcal{Y}_{t=2}^*$ &$\big\{ y^{*}_{2,j=1} {=} $ Buy$_{k=1}$ \textit{Entity}S1  \textit{Entity}U3 book$_{k=4}$,\; $y^{*}_{2,2} {=} $ Buy \textit{Entity}U3 book by \textit{Entity}S1 $\big\}$
\end{tabular}
  \caption{An example of sequential input received by our utterance disambiguation seq2seq model and a list of reference outputs. The words in short-caps denote the domain and intent predicted by the SLU system which are concatenated to the beginning of the sequence. Words beginning with \textit{Entity} are placeholders used to delexicalize names of entities. Both references 1 and 2 are input to the SLU system during training. We explicitly named the indices at a few locations to aid the reader.}
    \label{fig:data}
\end{figure*}

\textbf{Task} We define a sequence of $D$ dialogue turns, $\mathbf{x_t} = (u_{t-D+1}, r_{t-D+1}, \ldots, u_{t-1}, r_{t-1}, u_t)$, where $u_t$  is the user utterance at time $t$ and $r_t$ is the corresponding system response. $\mathbf{x_t}$ is the total information that our system has at time $t$. For example, the first row in \figref{fig:data} shows $\mathbf{x_2}$ encoded as a single token sequence corresponding to the dialogue in \figref{fig:dialog1}. The query rewriting task is to learn a function $f_\theta$,  with parameters $\theta$, which maps $\mathbf{x_t}$ to its \textit{rewrite} $\mathbf{y_t}$ which is another string, i.e. $\mathbf{y_t} = f_\theta(\mathbf{x_t})$. $\mathbf{y}_t$ should contain all the information needed by the agent to fulfill the user's request and it should be understandable by the agent as a standalone user request.

\textbf{Model} We use the pointer-generator (PGN) architecture~\cite{see2017gttp} to construct $f_\theta$. The PGN is a hybrid architecture which combines sequence-to-sequence model with pointer networks. This combination allows the PGN to summarize an input sequence by either \textit{copying} from the input sentence, or \textit{generating} a new word with a decoder RNN. We now describe the operation of the PGN in detail and focus on a single input sequence $\x$ with the subscript $t$ omitted for simplicity. Let us slightly abuse notation and consider $\x, \y$ as sequences of tokens. We index the tokens of $\x, \y$ by $l, k$ respectively. The PGN uses a two-layer Bi-Directional LSTM (BiLSTM) encoder to compute the hidden state vector $h_l$ for $\x_l$.\footnote{For sake of brevity, we omit the update equations for the LSTM. Please refer to \cite{see2017gttp} for these details.}

%We chose to use the PGN because it can summarize an input sequence by either  The decision between copying and generation is also dynamically predicted by the model at each time step. Because a large number of tokens in $\mathbf{y_t}$ also occur in $\mathbf{x_t}$ therefore the PGN is well suited for utterance rewriting.

 %We slightly abuse notation and consider $\x, \y$ as sequences of tokens from a common vocabulary, say $\mathcal{V}$.   The PGN architecture uses a $2$-layer Bi-LSTM RNN to compute the hidden state vector $h_l$ for $x_l$.  The PGN's decoder is a single-layer LSTM RNN. At decoding time $k$, the PGN's decoder computes three outputs: $p^{\text{copy}}$: A copy distribution over $\x$, $p^{\text{gen}}$: A generation distribution over $\mathcal{V}$, and $p^{\text{mix}}$: a mixture coefficient that is used to interpolate the copy and generating distribution.

We now describe how $y_k$ is generated.  At time $k$, the probability of copying a token from the input  $p^{\text{copy}}$ is computed via a softmax over the attention weights computed using non-linear function of the encoder-LSTM hidden states $\bh$ and the decoder LSTM's hidden state $h^{\text{decoder}}_k$. $p^{\text{mix}}$ --  a soft switch to decide between copying and generating -- is computed using another non-linear function of $h^{\text{dec}}_k$ and the final output distribution is given by
\begin{equation}\label{eq:mix_prob}
p(y_k) = p^{\text{mix}} p^{\text{gen}}(y_k) + (1 - p^{\text{mix}}) p^{\text{copy}}(y_k)
\end{equation}
At decoding time, we can use either beam-search or greedily pick the token with the highest probability and move on to the next step. This is our baseline architecture for utterance rewriting.

\textbf{Evaluation} Ideally $\mathbf{y_t}$ should be judged as a correct rewrite if the downstream SLU system can parse $\mathbf{y_t}$, invoke the correct agent with the correct slots, and the agent can then take the right action. However, evaluating this notion of correctness would have required probing and instrumenting thousands of downstream agents and is not scalable to implement. Therefore, we used a simpler notion of correctness based on a manually collected set of \textit{golden rewrites}, $\mathcal{Y}^*_{i,t}$, in this paper. \secref{ssec:metrics} describes the metrics we use to evaluate our model's prediction $y_{i,t}$ against the golden set $\mathcal{Y}^*_{i,t}$.

\textbf{Learning}  For training the model, we have a rewrites-corpus $\{\x_{it},\y^{*}_{itj}\}_{i=1,t=1,j=1}^{I,T,J}$. $I$ is the number of dialogs, $T$ is the maximum number of turns in a dialog and $J$ is the number of gold rewrites at a turn in a dialog. $\y^{*}_{i,t,j}$ denotes the $\ath{j}$ optimal rewrite for the user utterance at turn $t$ in the $\ath{i}$ dialogue -- $\x_{ti}$; $y^*_{i,t,j,k}$ is the $\ath{k}$ token in $\mathbf{y^*_{i,t,j}}$. Our training objective is to maximize the log-likelihood:
\begin{equation}\label{eq:ll}
\arg\max_{\theta} \sum_{i,t,j,k} \log p_\theta(y^*_{i,t,j,k}).
\end{equation}

\subsection{Multi Task Learning (MTL): Entity-Copy Auxiliary Objective}
\label{sec:mtl}
\begin{figure*}[h!]
  \centering
 \includegraphics[width=\linewidth]{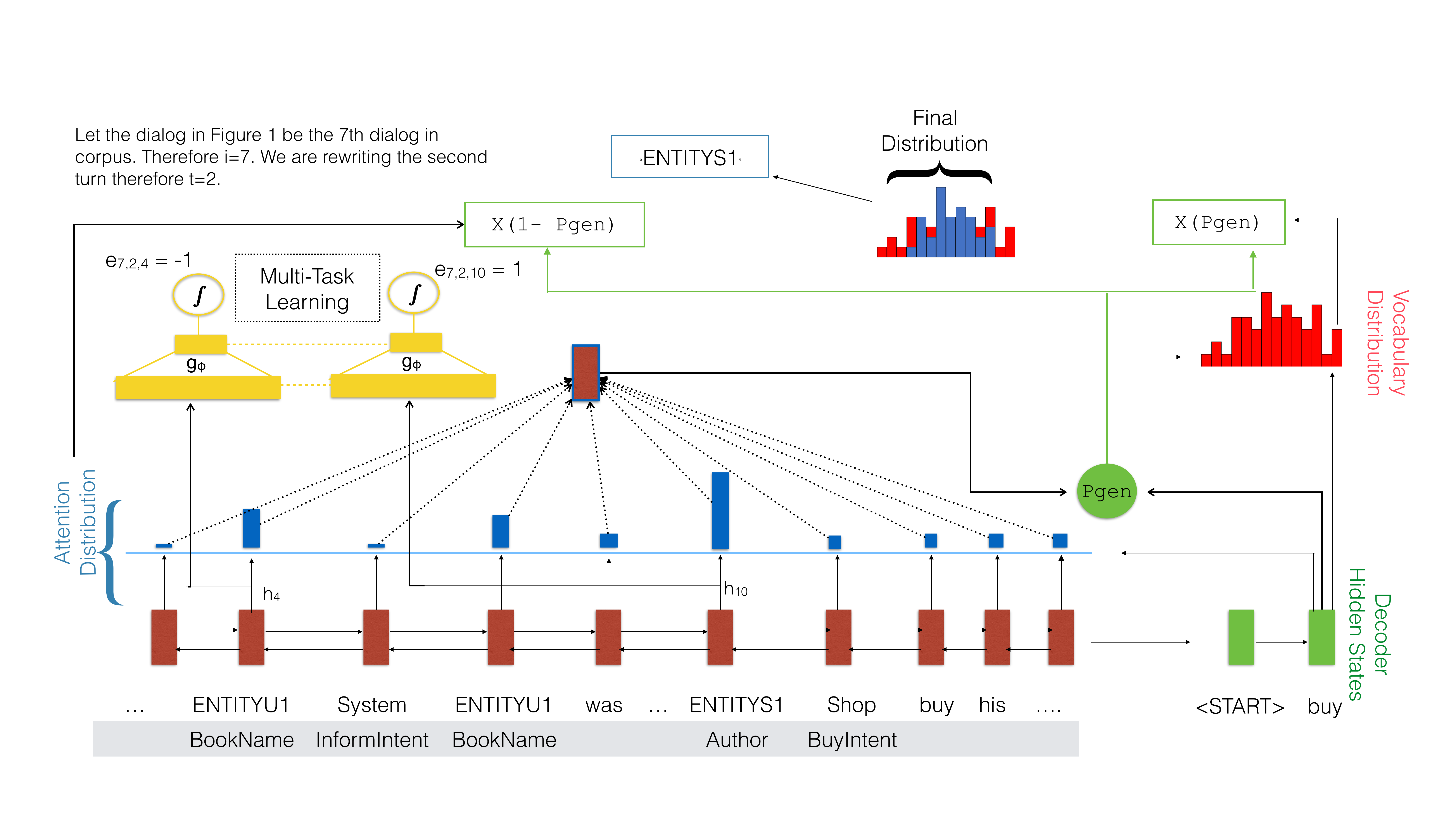}
  \caption{Model Architecture of the CQR Model which performs Multi-Task Learning for Pointer-Generator Networks. We show a snapshot just before decoder generates the word {\textsc{EntityS1}} or \textit{Yuval Harari}. Also, we show for MTL {\textsc{EntityU1}} gets the label $-1$ as it is not one of the final slots, and {\textsc{EntityS1}} gets a label of 1}
    \label{fig:arch}
\end{figure*}

In~\figref{fig:data}, both the references $y_{2,1}^{*}, y_{2,2}^{*}$ contain the same subset of entities -- U3, and S1 -- even though their order, and other tokens, in the gold rewrites have changed. This implies that for the task of rewriting utterances, the subset of entities that should be copied from the input dialog remains the same, irrespective of the dynamics of the decoder LSTM. Based on this observation we define an auxiliary task and augment the learning objective as shown in~\figref{fig:arch}.

As mentioned earlier, the copy distribution $p^{\text{copy}}_k$ is a function of the encoder hidden state  $\bh = (h_1, \ldots, h_l, \ldots, h_{|x|})$ which does not change with $k$. If $x_l$ was an entity token then $h_l$ should be informative enough to decide whether that token should be copied or not. Therefore, we add a two layer feed-forward neural network, $g_\phi$, that takes $h_l$ as input and predicts whether the $\ath{l}$ token should be copied or not. Given the probability $g_\phi(h_l)$ we minimize the binary cross-entropy loss, and back-propagate through $h_l$ which influences $\theta$. The auxiliary objective should improve the generalization because it forces the encoder’s representation to become more informative about whether an entity should be copied or not. At inference time $g_\phi$ is not used. Formally, let $e_{i,t,l}$ take the following value:
\begin{equation*}\label{eq:mtl_desc}
  e_{i,t,l} = \begin{cases}
    \hphantom{-}1 &\text{if $x_{i,t,l}$ is an entity and } x_{i,t,l} \in \mathcal{Y}^*_{i,t}\\
    -1 &\text{if $x_{i,t,l}$ is an entity and } x_{i,t,l} \notin \mathcal{Y}^*_{i,t}\\
    \hphantom{-}0 & \text{Otherwise}
  \end{cases}
\end{equation*}
Let $\lambda > 0$ be a hyperparameter. We add a binary log-likelihood objective to objective~\ref{eq:ll} to create objective~\ref{eq:mtl}. We refer to the PGN model trained with objective~\ref{eq:mtl} as \textbf{CQR} in \tabref{tab:ourdata}.
\begin{equation}\label{eq:mtl}
  \kern-10pt{\sum_{i,t,j,k}} \log p(y^*_{i,t,j,k}) {+} \lambda \sum_{i,t}\kern-2pt\sum_{l=1}^{|x_{i,t}|} \kern-3pt e_{i,t,l} \log g_{{\phi}}(h_{i,t,l})
\end{equation}

%% file: data.tex
In this section we will describe how we created the golden rewrites $\{\mathcal{Y}_t^* \mid \forall t\}$ for each of the above datasets and our pre-processing steps that we found crucial to our success.

\subsection{Generating gold rewrites}
\label{sec:gen_rewrites}
We used two separate approaches to generate gold rewrites for the \textsc{Internal} and \textsc{InCar} datasets.
For the \textsc{InCar} dataset we collected 6 rewrites for each utterance that had a reference to a previously mentioned entity.\footnote{We will release these annotations as an extension to \cite{eric2017key}.} 
For the \textsc{Internal}  dataset, which has over 100K sentences the above approach would be prohibitively expensive.
Therefore, instead of gathering completely new manual annotation we used a semi-manual process.
We utilized a template dataset that is keyed by the \textit{Domain, Intent} and \textit{Slots} present in that utterance and contains the top-5 most common and unambiguous phrasing for that key.
For example to create the rewrite in Figure 1 we filled the template:
\[
\text{{Buy} \underline{\color{blue}Creator} 's \underline{\color{blue}SortType} \underline{\color{blue}ItemType}}
\]
This template was chosen randomly from other valid alternatives such as {{Buy} {\color{blue}SortType} {\color{blue}ItemType} by {\color{blue}Creator} }. These valid alternatives were determined on the basis of existing manual \textit{domain, intent,} and \textit{schema} SLU annotations which indicated which slots were required to answer the user's utterance.

\subsection{Role-based Entity Indexing}
\label{sec:rbi}
In this step, the entity words in $x_t$ are replaced with their canonical versions. Our results show that this significantly improved both BLEU and Entity F1 measures. To replace entity words we use string matching methods to extract tokens for dialogue. We maintain two separate namespaces for user entities and system entities respectively. However, if an entity appears again in dialogue, we do \textbf{\textit{not}} assign it a new canonical token but used already assigned one. Also, as seen in Figure 2 we also add the entity tag to slot representation. Lastly, as re-writing happens before any SLU component we do not have this information for $u_t$. In $u_t$ we only replace entities with canonical tokens, but do not add any information about entity. \tabref{tab:ppp1} show how to transform dialogue from Figure 1.

\begin{table}[H]
\centering
%\resizebox{\columnwidth}{!}{
\begin{tabular}{|p{3.5cm} |p{3.5cm}|}\hline
\textbf{Before} & \textbf{After Pre-Processing} \\\hline
\small Who wrote Sapiens & \small who wrote U\_1$||$BookName  \\\hline
\small Sapiens was written by Yuval Harari& \small U\_1$||$Author was written by S\_1$||$BookName\\\hline
\small Buy his most recent book& \small Buy U\_3$||$UNK U\_4$||$UNK book  \\\hline
\end{tabular}
%}
\caption{Replacing entities with the role-based canonical versions.}
\label{tab:ppp1}
\end{table}

\subsection{Abstractified Possessives}
\label{sec:abstr-poss}
Generalizing on rare words and rare contexts is the true test of any NLP system, and linguists have long argued in favor of syntactically motivated models that abstract away from  lexical entries as much as possible~\cite{klein2003}. In this preprocessing step, we show the benefit of such abstraction. While testing the PGN architecture we noticed that the sequence decoder would sometimes generate an off-topic rewrite if the input sequence contained a rare word. In order to avoid this problem we augmented the input sequence with additional features to mark the syntactic function of words. Specifically we used the Google Syntactic N-gram Corpus \cite{goldberg2013dataset} to add syntactic features to each word in the dialogue. We harvested a list of top $1000$ words that appear most frequently after possessive pronouns.  We concatenated three types of extra features to the words in a dialogue. The first feature was the \textsc{Question} feature which was concatenated to the $7$ question words. The second feature was the $PRP\$$ tag which we concatenated to specific possessive pronouns. Finally we added a tag called \textsc{PSBL} -- short for possessible -- for the top $1000$ words that we found from the Syntactic $N$-Gram Corpus. 
%\begin{table}[H]
%\centering
%\begin{tabular}{|c | c|}\hline
%Before & After pre-processing \\\hline
%who is his wife & who$\vert$Question is  \\
%  & his$\vert$PRP\$ wife$\vert$PSBL  \\\hline
%\end{tabular}
%\caption{Adding syntactic features to $x_t$}
%\label{tab:ppp2}
%\end{table}

We decided not to use POS tags because we did not have manually POS tagged data on our domain and off-the-shelf POS tagger\footnote{https://spacy.io/} did not perform well on our dataset.

%% file: experiments.tex
\subsection{Dataset}
We used two datasets to evaluate our method. The first is a public dataset ~\cite{cqrdataset} we call \textsc{InCar}, which is  an extension to \cite{eric2017key}.  The dataset consists of $3,031$ dialogues from three domains: Calendar Scheduling, Weather, and Navigation, that are useful for an in-car conversational assistant. We crowd-sourced six rewrites for each utterance in the corpus that had a reference to previously mentioned entities. The second dataset, called \textsc{Internal}, is an internal benchmark dataset we collected over six domains -- weather, music, video, local business search, movie showtimes and general question answering.~\tabref{tab:dss} describes the data statistics for this internal collection. About $40\%$ of the dialogues in this corpus are cross-domain, which makes it much harder than the \textsc{InCar} dataset.

%\begin{table*}[htbp]
%  \centering
%  \begin{tabular}{|c | ccc  | ccc | ccc |}\hline
% &  \multicolumn{3}{|c|}{Train} &  \multicolumn{3}{|c|}{Dev} &  \multicolumn{3}{|c|}{Test}\\
%Domain & 1 & 2 & $\geq$3 & 1 & 2 & $>=$3 & 1 & 2 & $>=$3 \\\hline
%Local-Business & 12K & $\sim$400 &  $\sim$150 & 4K & 120 & 60  & 2K & 60 & 10 \\\hline
%Music & 12K & 4K & 2K & 4K & 1.5K & 700 & 2K & 750 & 350 \\\hline
%Weather & 10K & 2K & 1K & 3.6K & 600 & 300 & 2K & 300 & 150 \\\hline
%Q\&A & 10K & - & - & 3.6K & - & - & 18K & - & - \\\hline
%Video & 4K &  $\sim$100 & 30 & 1K & 40 & 10 & 500 & 20 & 5\\\hline
%Movie-Showtimes & 20K & 50 & 20 & 6K & - & 10 & 3K & 20 & - \\\hline
%Books & 1K & 200 & 100 & 417 & 70 & 30 & 200 & 35 & 20 \\\hline
%Others & 2K & 700 & 400 & 700 & 200 & 125 & 300 & 100 & 80 \\\hline
%Cross-Domain & 55K & 300 & 200 & 18K & 100 & 100 & 9K & 50 & 40 \\\hline
%Total & 125K & 8K & 4K & 42K & 3K & 1K & 21K & 1K & 700 \\\hline
%    \end{tabular}
%
%  \caption{\small Internal Data statistics for train, dev, test partitions. Dataset contains 8 large domains, and few others with small representation. Unlike, \textsc{InCar} dataset, there is a significant portion of dataset which consists of utterances from multiple domains. We show statistics for context length $1, 2$, and $\geq$ 3, each turn consist of a user and a system turn, i.e. a carried slot maybe from as far as 6 utterances in the dialog.}
%  \label{tab:dss}
%\end{table*}

\begin{table}[htbp]
  \centering
  \begin{tabular}{|c|c|c|c|}\hline
Context Length & Train & Dev & Test \\\hline
1 & 125K & 42K & 21K \\\hline
2 & 8K & 3k & 1K  \\\hline
$>=$3 & 4K & 1K & 700 \\\hline
\end{tabular}
  \caption{\small \textsc{Internal} data statistics. Each turn consists of a user and a system turn i.e context length = 2 implies two turns.}
  \label{tab:dss}
\end{table}

\subsection{Training}
We used OpenNMT \cite{opennmt} toolkit for all our experiments. We modified it to include the multi-task loss function as described in~\secref{sec:mtl}. Unless explicitly mentioned here, we used the default parameters defined in OpenNMT recipe. Various hyper-parameters were tuned on a reduced training set and the development set. Our encoder was a 128-dimensional bi-directional LSTM. We used the Adagrad optimizer with a learning rate of $0.15$, and we randomly initialized 128-dimensional word embeddings. The word embeddings were shared between the encoder LSTM and the decoder LSTM. $\lambda$ in Eq.\ref{eq:mtl} was set to $0.01$. We trained the model for $20$ epochs with early stopping on a validation set.

\subsection{Evaluation Metrics}
\label{ssec:metrics}

\textbf{BLEU:} has been widely used in machine translation tasks \cite{papineni2002bleu},  dialogue tasks \cite{eric2017key}, and chatbots \cite{ritter2011data}. It gives us an intrinsic measure to evaluate quality of re-writes without caring about downstream SLU evaluation.

\textbf{Response Entity F1 (ResF1):} We measure this metric for the \textsc{InCar} dataset, following the approach outlined by \cite{mem2seq2018}\footnote{Evaluation script available at https://github.com/HLTCHKUST/Mem2Seq}. The Response Entity F1 micro-averages over the entire set of system responses and compare the entities in plain text. The entities in each gold system response are selected by a predefined entity list. This metric evaluates the ability to generate relevant entities and to capture the semantics of the dialogue.  We reimplemented the \textit{Mem2SeqH1} architecture in~\cite{mem2seq2018}\footnote{The \textit{Mem2Seq H1} was the best performing system in terms of ResF1, in two out of three domains in the InCar dataset, and it was the fastest Mem2Seq model. Therefore, we used \textit{Mem2SeqH1} and not \textit{Mem2SeqH3}} and we refer to our implementation as \textit{Mem2Seq}$^*$. We use utterances produced by our proposed \textbf{(CQR)} system in the dialogue instead of original utterances while evaluating using Mem2Seq$^*$. Note that our reimplementation, Mem2Seq$^*$, achieves a Response Entity F1 of $33.6$ which is higher than the best overall Entity F1 score of $33.4$ reported in~\cite{mem2seq2018}.

\textbf{Entity F1}: This measures micro F1 between entities in the hypothesized rewrite and gold rewrite. This is different from F1 reported by \cite{mem2seq2018} as they evaluate F1 over system entities, whereas here we evaluate the entities over the user turn.  We employ a recent state-of-art bi-directional LSTM with CRF decoding~\cite{ma2016end} to implement our SLU system.

%% file: results.tex
\subsection{\textsc{Internal} Dataset Results}
On \textsc{Internal} dataset we show CQR significantly improves over \cite{naik2018contextual} in \tabref{tab:ourdata}. \textbf{CQR} also improves F1 for current turn slots as it can leverage context and distill necessary information to improve SLU. Further, we can see that most improvements upon the baseline PGN model (M0) come from pre-processing steps like canonicalizing entities. In the baseline model, it has to learn to generate entity tokens individually, whereas in \textbf{M1} the model only has to learn to copy tokens like \textit{USER\_ENT\_1}. Finally, our proposed multi-task learning model (CQR) improves both BLEU and EntityF1 at most distances. Specifically, we see an improvement of 4.2\% over \textbf{M2} for slots at distances $\geq$3. In \tabref{tab:ourdata} distance is measured differently from \tabref{tab:dss}, here we count User and System turns individually to showcase how distance affects \textbf{EntityF1}. If an entity is repeated multiple times in the context, we consider its closest occurrence to report results.

\begin{table}
\small
\begin{tabular}{| p{1.2cm}|p{5.6cm}|}\hline
\textbf{Dialogue} & U: Find me a Starbucks in Redmond \\
& S: I found a Starbucks in Redmond WA. It's 15.7 miles away on NE 76th St. It's open now until 9:00 PM. \\
& U: How do I get there? \\\hline
\textbf{PGN} & how can i get to redmond \\
\textbf{CQR} & how do i get to the starbucks on NE 76th St WA \\
\textbf{Gold} & how do i get to the starbucks on NE 76th St \\\hline\hline
\textbf{Dialogue} & U: How is the weather tomorrow? \\
& S: In Chicago there will be mostly sunny weather \\
& U: What about saturday? \\\hline
\textbf{PGN} & what is the weather in chicago on saturday ? \\
\textbf{CQR} & what is the weather in chicago on saturday ? \\
\textbf{Gold} & what is the weather in chicago on saturday ? \\\hline
\end{tabular}
\caption{Examples of generated responses for Internal Dataset}
\label{tab:rweg}
\end{table}

\begin{table*}[htbp]
%\resizebox{\columnwidth}{!}{
  \centering
 \begin{tabular}{|p{3.5cm} | p{5mm}p{6mm}p{5mm} |p{5mm}p{5mm}p{5mm} |  p{5mm}p{5mm}p{5mm} |p{5mm}p{5mm}p{7mm} | p{9mm}| }\hline
   \multicolumn{1}{|r}{\textbf{System}} &  \multicolumn{12}{|c|}{\textbf{Entity F1}}      &  \multicolumn{1}{|p{9mm}|}{\hskip-5pt\textbf{BLEU} }\\\hline
 &  \multicolumn{3}{|c|}{d$=$0}  &\multicolumn{3}{|c|}{d$=$1} & \multicolumn{3}{|c|}{d$=$2}  & \multicolumn{3}{|c|}{d$\geq$3} & \\
& P & R & F1  & P & R & F1   &  P & R & F1 &  P & R & F1 &  \\\hline
 \multicolumn{1}{|r|} { \cite{naik2018contextual}} & 95.1&\textbf{95.1}&95.1&74.9&78.4&76.6&72.2&82.2&76.9&10.4&46.3&17.0& \multicolumn{1}{|c|}{N/A}           \\\hline
 \multicolumn{1}{|r|} {PGN (M0)}&\textbf{99.0}&78.1&87.4&\textbf{95.9}&62.1&75.4&\textbf{95.2}&57.3&71.5&87.3&65.5&74.9& \multicolumn{1}{|c|}{83.4}\\\hline
 \multicolumn{1}{|r|}  {${+}$Canonical Ent. (M1)}&98.7&93.9&\textbf{96.3}&92.9&93.5&93.2&94.4&96.9&95.6&69.8&78.5&73.9& \multicolumn{1}{|c|}{89.9}  \\\hline
  \multicolumn{1}{|r|} {${+}$Syntax Info (M2)}       &98.6&93.9&96.2&92.9&93.5&93.2&94.3&96.9&95.6&69.8&78.5&73.9&\multicolumn{1}{|c|}{89.9} \\\hline
  \multicolumn{1}{|r|} {${+}$MTL (CQR)}                    &98.5&\textbf{94.0}&96.2&93.7&\textbf{93.8}&\textbf{93.7}&94.2&\textbf{97.4}&\textbf{95.8}&\textbf{75.2}&\textbf{79.0}&\textbf{77.1}&\multicolumn{1}{|c|}{\textbf{90.3}}  \\\hline
  \% Relative Improv. & 3.6	&-1.2	&1.2	&25.1	&19.6	&22.3	&30.5	&18.5	&24.6	&623.1	&70.6	&353.5&N/A\\\hline
\end{tabular}
%}
\caption{Comparison of Pointer-Generator variants to traditional state tracking approach on the \textsc{Internal} dataset. We measure entity F1 across slots from different distances separately. Slot distance is counted per utterance starting from the current user utterance. Therefore, slots at d$=$0 are slots from the current user utterance that should have been copied. d$=$1 refers to slots from system response in the last turn, d$=$2 refers to slots from the user in last turn and d$\ge$3 aggregates all other turns. $d\ge$3 is the most challenging test-subset where \textbf{CQR} has the highest benefit.}
\label{tab:ourdata}
\end{table*}

\subsection{\textsc{InCar} Dataset Results}
For \textsc{InCar} dataset we pick the best model \textbf{CQR} from \tabref{tab:ourdata} and re-train on the respective dataset. On the navigation domain we observe significant improvement. We believe this is because there are on average $2.3$ slots were referred from history in rewrites requiring copy from dialog as compared to $1.3$ and $1.1$ in schedule and weather domain respectively. Also, we compare with an oracle CQR (i.e., gold-rewrite from our data collection, instead of predicted re-write) to measure the potential of query-rewriting and motivate further research on this topic. We can see that the CQR model performs better than the Mem2Seq$^*$ model, indicating that query rewriting is a viable alternative to dialogue state tracking. This is important in environments where changing the NLU systems to leverage memory structures is not always feasible. We claim that query rewriting is a simpler approach in such situations, with no loss in performance.

\begin{table*}[h!]
\centering
%\small
\setlength{\tabcolsep}{4pt}
\begin{tabular}{| r | c | cccc |}\hline
 \textbf{System} & \textbf{E2E} & \multicolumn{4}{|c|}{\textbf{ResF1}} \\
&  \textbf{BLEU}    & All & Schedule & Weather & Navigation \\\hline
%Mem2Seq~\cite{mem2seq2018}       & $11.6$ & $32.4 $ & $39.8$ &$33.6$ & $24.6$   \\
Mem2Seq$^*$     & $11.4$ & $33.6 $ & $\textbf{48.4}$ &$47.2$ & $19.4$      \\\hline
\textbf{CQR}& $\textbf{11.6}$ & $\textbf{36.1}$ & $\textbf{48.4 }$ & $\textbf{47.9}$ & $\textbf{23.8}$        \\\hline
\% Relative Improv. & $1.8$ &$7.4$ &$0.0$ &$1.5$ &$22.7$\\\hline\hline
CQR-Oracle         & $11.8$ & $38.0 $ & $48.9$ &$48.9$ & $26.9$ \\\hline

\end{tabular}
\caption{Comparison of PGN variants proposed in this paper on the \textsc{InCar} dataset in comparison to the state tracking approach. Our proposed CQR model outperforms the MemSeq* system, which is a stronger baseline than the Mem2Seq results published in~\citet{mem2seq2018}.}
\label{tab:incar}
\end{table*}

%% file: background.tex
Probabilistic methods for task-oriented dialogue systems typically divide  an automatic dialogue agent into modules such as automatic speech
recognition(ASR) for converting speech to text, spoken language understanding(SLU) for classifying the domain and intent of the current utterance and tagging the slots in the current
utterance, dialogue state tracking(DST) for tracking  what has happened in the dialogue so far, and dialogue policy for deciding what actions to take next~\cite{young2000probabilistic}. In this traditional framework, SLU is seen as a low-level task that interprets the user's current utterance in isolation, without accounting for the dialogue history. For example, in \figref{fig:dialog1} the platform system parses the utterance
\textsc{Who wrote Sapiens},  to infer that the user intends to query for information about a book, and then the platform performs BIO style tagging with an intent-specific schema to label the mention\textsc{Sapiens} as the slot key \textsc{BookName}. Most SLU systems perform this without any context information. Some recent work focussed on contextual SLU~
\cite{contextual-slu-ms, liu2015deep, chen2016contextualslu} propose memory architectures to incorporate contextual information while performing the SLU step. However because their task was restricted to domain-intent classification and slot tagging for the current utterance only, a higher level DST module is still required to combine information from previous turns with the current utterance to create a single dialogue state.

DST is considered to be a higher-level module as it has to combine information from previous user utterances and system responses with the current utterance to infer its full meaning.
Many deep-learning based methods have recently been proposed for DST such as neural belief tracker~\cite{mrkvsic2017neural}, and self-attentive dialogue state tracker~\cite{zhong2018global} which are suitable for small-scale domain-specific dialogue systems; as well as more scalable approaches such as~\cite{rastogi2017scalable,xu2018e2e} that solve the problem of an infinite number of slot values and~\cite{naik2018contextual} who additionally solve the problem of huge number of disparate schemas in each domain. End-to-End approaches based on deep learning have also been proposed recently to replace such modular architectures, like~\cite{mem2seq2018,eric2017key}.

Unfortunately, all of the above approaches fail to address the problem that, as the number of domain-specific chatbots on a dialogue platform grows larger, the DST module becomes increasingly complex as it tries to handle the interactions between different chatbots and their different schemas. For example, consider the scenario shown in \figref{fig:dialog1}.
Chatbot A, the \textsc{Book} chatbot, can understand domain-specific utterances like ``who wrote X ?'' annotated with a special schema with slot keys such as \textsc{BookName, Author}. In order to disambiguate utterance $u_2$ the DST in the conversational platform must know that the \textsc{Creator} slot-key in the \textsc{Shopping} chatbot co-refers to the
\textsc{Author} slot-key. However, this leads to a quadratic explosion in the number of possible transitions that the platform has to learn, thereby significantly increasing the learning problem for DST. Additionally, the problem is more challenging than just disambiguating pronouns because in some situations there may be no co-referent pronouns in the current utterance. For example, a user may say ``what's the address'' instead of saying ``what is its address'', creating a case of zero-anaphora.

Finally, we will mention that Seq2Seq models with Attention~\cite{sutskever2014sequence,bahdanau2014neural} have seen rapid adoption in automatic summarisation~\cite{see2017gttp,rush2015}. Exploring black-box methods like query re-writing allow us to benefit from the progress made in these fields and apply them to state tracking and reference resolution tasks in dialogue.

%% file: conclusion.tex
In this work we made three fundamental contributions. First, we proposed {\it contextual query rewriting}(CQR) as a novel way to interpret an input utterance in context given a dialogue history. For example, we can rewrite \textsc{Buy his latest book} as \textsc{Buy Yuval Harari's most recent book}, given the dialogue history, as shown in \figref{fig:dialog1}. The output of CQR can directly be fed to the domain-specific downstream SLU system which drastically simplifies the construction of task-specific dialogue \textit{agents}.  Since we do not need to change either the spoken language understanding or the dialogue state tracker downstream, our approach is a black-box approach that improves the modularity of industrial-scale, dialogue-asistants. Second, we investigated how to optimally use a Pointer-Generator Network for the CWR task using Multi-Task Learning and task-specific preprocessing. Finally, we demonstrated the efficacy of our approach on two datasets. On \textsc{InCar} dataset released by~\cite{eric2017key}, we were able to show that re-writing of the user utterance can benefit end-to-end models. On a proprietary \textsc{Internal} dataset we showed that our approach can greatly improve the experience when referring to entities from much further away in a dialogue history, resulting in relative improvements in Entity F1 of greater than 20\% on the most challenging subset of the test-data.

We hope that our approach of directly using natural language as an api will motivate other researchers to conduct work in this direction.